\pdfoutput=1

\documentclass[11pt]{article}

\usepackage{authblk}
\usepackage[]{ACL2023}

\usepackage{times}
\usepackage{latexsym}
\usepackage[fleqn]{amsmath}
\usepackage{amsfonts}

\usepackage{mathtools}
\usepackage{bbm}

\usepackage[T1]{fontenc}

\usepackage[utf8]{inputenc}

\usepackage{microtype}

\usepackage{inconsolata}
\usepackage{paralist}
\usepackage{graphicx}
\usepackage{mdframed}

\usepackage{url}
\usepackage{xspace}

\def\seclabel#1{\label{sec:#1}\label{p:#1}}
\def\secref#1{Section~\ref{sec:#1}}

\def\impressions{\textsc{impressions}\xspace}
\def\findings{\textsc{findings}\xspace}
\def\clinicalsection{\textsc{clinical section}\xspace}
\def\comparison{\textsc{comparison}\xspace}
\def\miscellaneous{\textsc{miscellaneous}\xspace}

\def\Electra{\textsc{electra}\xspace}

\def\figlabel#1{\label{fig:#1}\label{p:#1}}
\def\figref#1{Figure~\ref{fig:#1}}

\title{RadLing: Towards Efficient Radiology Report Understanding}

\author[1]{\bf Rikhiya Ghosh}
\author[1]{\bf Sanjeev Kumar Karn}
\author[2]{\bf Manuela Daniela Danu}
\author[2]{\bf Larisa Micu}
\author[ ]{\authorcr \bf Ramya Vunikili}
\author[1]{\bf Oladimeji Farri}
\affil[1]{Digital Technology and Innovation, Siemens Healthineers, Princeton}
\affil[2]{Corporate Technology, Siemens AG, Romania}
\affil[ ]{\tt \{rikhiya.ghosh,sanjeev.kumar\_karn,oladimeji.farri\}@siemens-healthineers.com}
\affil[ ]{\tt \{manuela.danu,larisa.micu\}@siemens.com}

\begin{document}
\maketitle
\begin{abstract}
Most natural language tasks in the radiology domain use language models pre-trained on biomedical corpus. There are few pretrained language models trained specifically for radiology, and fewer still that have been trained in a low data setting and gone on to produce comparable results in fine-tuning tasks. We present RadLing, a continuously pretrained language model using \Electra-small \citep{clark2020electra} architecture, trained using over 500K radiology reports, that can compete with state-of-the-art results for fine tuning tasks in radiology domain. Our main contribution in this paper is knowledge-aware masking which is a taxonomic knowledge-assisted pretraining task that dynamically masks tokens to inject knowledge during pretraining. In addition, we also introduce an knowledge base-aided vocabulary extension to adapt the general tokenization vocabulary to radiology domain. 
\end{abstract}

\section{Introduction}
Radiology reports are radiologist interpretations of medical images such as X-Rays, CT, Ultrasound and MRI scans. Healthcare professionals rely on these reports to monitor and diagnose patients. A radiology report typically includes several sections \citep{kahn2009toward}, among which the most important ones are the following: 
\begin{enumerate}
    \item \clinicalsection. This section describes afflictions of the patient that prompted the study, past diseases and symptoms.
    \item \comparison. This refers to previous imaging studies of the patient with which the radiologist is comparing the current image.
    \item \findings. This section includes qualitative and quantitative descriptions of abnormalities if present, along with the radiologist's diagnosis or differential diagnosis regarding the observations.
    \item \impressions. This section summarises the \findings section. The radiologist notes major abnormalities and their recommendations.
    \item \miscellaneous. This consists of other information like patient information, imaging modality. 
\end{enumerate}
The post-BERT era \citep{devlin2018bert} of contextualized pretrained language models (PLMs) has drastically reduced the need for expensive and hard-to-find human annotated data for biomedical NLP. Biomedical PLMs are usually trained on biomedical publications \citep{pubmedbert, lee2020biobert, gururangan2020don, peng2019transfer, alsentzer2019publicly, lin2021entitybert, yuan2022biobart, luo2022biogpt} and have facilitated drug discovery and healthcare informatics. Despite sharing the same concepts, challenges remain in adapting these models to radiology reports. This is because the contents, context and structure of biomedical publications are significantly different from those of the radiology reports. Radiology reports are terse, and concept-dense. It is shown that models pretrained on radiology reports improve performance in downstream clinical NLP tasks as opposed to models pretrained on biomedical publications \cite{yan2022radbert, smit2020chexbert, dai2021bonebert}. While this is encouraging, we identified some research challenges as well as encountered issues in adapting these PLMs to industrial setting:

\begin{figure}
\includegraphics[width=0.48\textwidth]{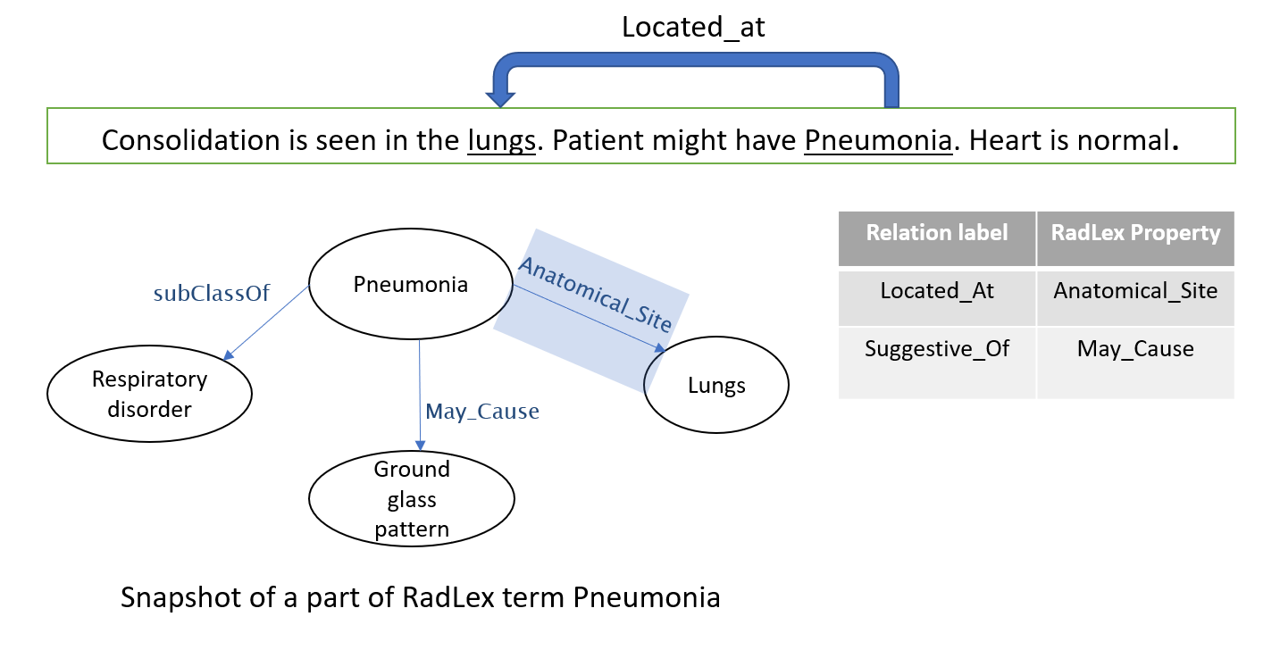}
\centering
\caption{A difficult relationship extraction problem in radiology report, taken from RadGraph relationship extraction dataset \cite{jain2021radgraph}. The shaded arrow shows corresponding RadLex entities being connected via the property \textit{Anatomical\_site}.}
\label{fig:ex}
\end{figure}

\noindent \textbf{Research challenges.} Random masking in masked language modeling (MLM) is known to have context understanding issues\citep{song2020mpnet}. This is exemplified in tasks like relationship extraction \citep{jain2021radgraph} in radiology reports. Figure \ref{fig:ex} shows an example sentence from CheXpert \citep{irvin2019chexpert}. The entities `Pneumonia' and `lungs' are related concepts, but current state-of-the-art PLM-based methods fail to identify their relation in the sentence.

\noindent \textbf{Industry adaptation issues.} Industry adaptation issues emanate from data and model sizes. Training datasets available for industry are generally small. Use of public datasets like MIMIC-IV \citep{johnson2020mimic} is not feasible for industry research due to license restrictions. Large PLMs like PubMedBERT (109 million parameters) and best performing RadBERT variant(125 million) have latency issues in industrial deployment, with low throughput when deployed in low memory settings. Reduction in parameters by quantisation or distillation reduces efficiency which is less than ideal.

We believe that context understanding problem can be avoided with the help of domain knowledge. In radiology domain, an excellent source of domain knowledge is the taxonomical knowledge base RadLex \citep{langlotz2006radlex}, which curates radiology lexicons. For the sentence in Figure \ref{fig:ex}, current PLMs misclassify the relationship. However, a look at RadLex for the entity 'Pneumonia' reveals its \textit{Anatomical\_site} property to be 'Lungs'. The RadLex property \textit{Anatomical\_site} and the RadGraph relation `Located\_at' are analogous. Infusing this knowledge into PLMs adds more context and domain knowledge and thus increases prediction capabilities. In this paper, we introduce RadLing, the first radiology language model based on \Electra-small architecture with 13.7 million parameters trained with the help of RadLex. Our major contributions in this paper are:
\begin{enumerate}[(a)]
    \item Domain-specific vocabulary extension where we have modified existing tokenization methods with help of RadLex,
    \item Knowledge base-aided continuous pretraining objective that abets better context understanding, and
    \item Smaller high performance radiology language model.
\end{enumerate}

\begin{figure*}[h!]
\includegraphics[width=0.95\textwidth]{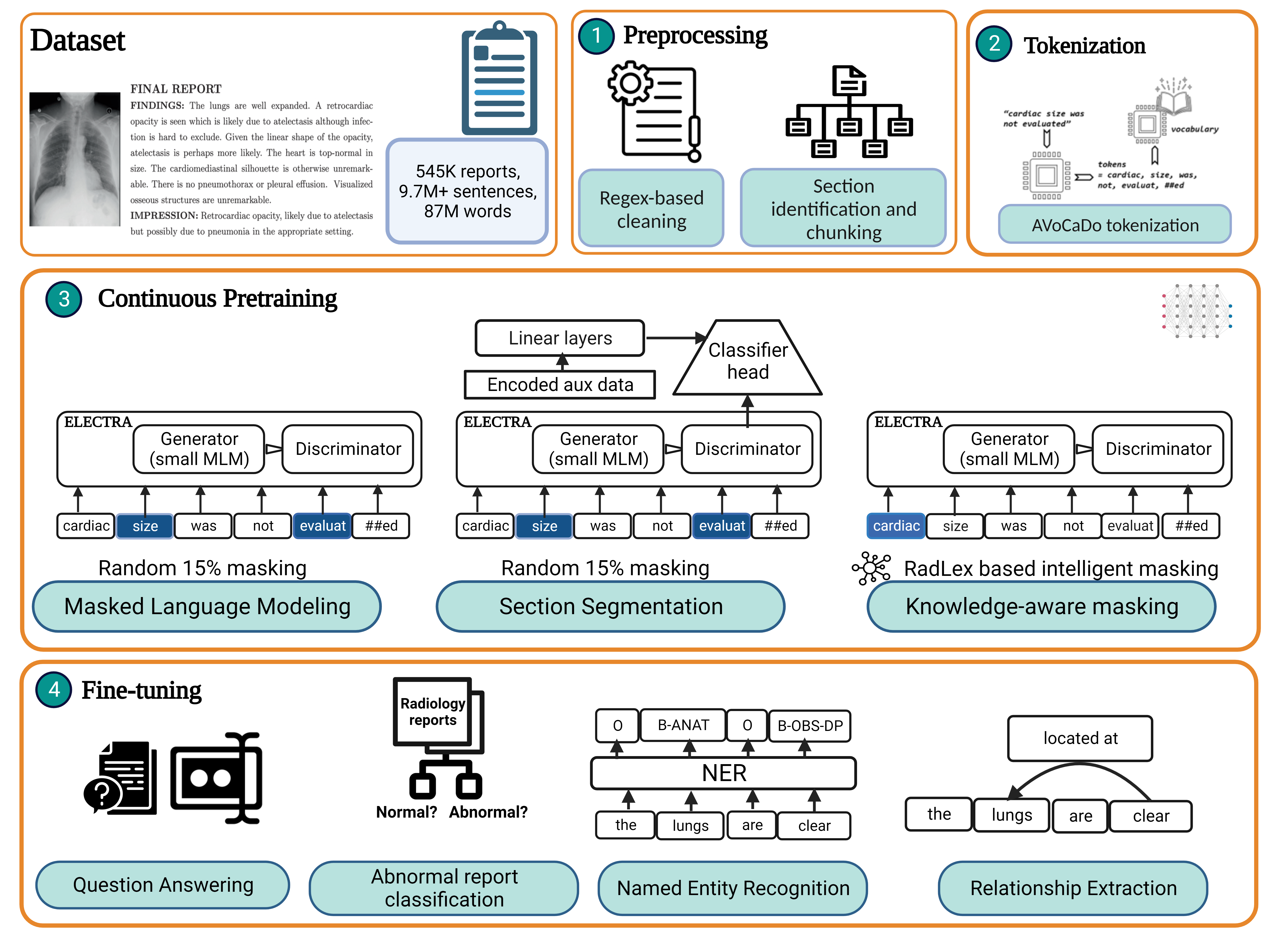}
\centering
\caption{Schematic details of the methods outlined in this paper. We perform Preprocessing, followed by tokenization, continuous pretraining and fine tuning. For continuous pretraining, this figure depicts the architecture used for the three pretraining objectives: Masked Language modeling, Section segmentation and Knowledge-aware masking. For fine-tuning, we have four tasks: Question Answering, Abnormal report classification, Named Entity Recognition and Relationship Extraction. Radiology report picture courtesy: \cite{liang2022development} Created with BioRender.com} 
\label{fig:schematic}
\end{figure*}

\section{Related Work}
Radiology Report Understanding and Radiology summarization are the most-addressed NLP tasks that are related to the field of radiology. Radiology summarization involves generating a summary of a radiology report, which can help clinicians quickly understand the key findings without having to read the entire report \cite{karn-etal-2022-differentiable}. Radiology Report Understanding involves extracting information from a radiology report, such as the type of imaging study, problem list generation, etc. This information can then be used to assist with diagnosis and treatment planning.

 \noindent \textbf{Radiology report understanding using MLMs.} \citet{yan2022radbert} has used MLM with several pretrained PLMs like BERT, RoBERTa\cite{liu2019roberta} and Clinical BERT \cite{huang2019clinicalbert} to train on 4.4 million radiology reports. These models show high performance in NER, RE, QA, abnormal classification and summarisation. In contrast, we aim to train with knowledge graph pretraining objectives that might help learn with much smaller dataset with 545k reports, and achieve comparable results for NER, RE and Abnormals classification. 
 
 \noindent \textbf{Other biomedical pretraining objectives.} EntityBERT \cite{lin2021entitybert} used entity masking pretraining objective to infuse domain knowledge. This pretraining objective can mask context in radiology reports, and hence we improved on this by using RadLex. Section segmentation is another radiology-specific pretraining objective proposed in  BiRadsBERT \cite{kuling2021bi} which we have adapted to our work RadLing-SS. We plan to use fewer sections in keeping with our training dataset structure and test the architecture for our training data.
 
\noindent \textbf{Biomedical Knowledge Graph infusion.} KeBioLM \citep{yuan-etal-2021-improving} applies text-only encoding layer to aggregate entity representation. They use MLM, entity detection and entity linking as pretraining objectives and surpassed state-of-the-art in biomedical NER and RE. BioKGLM\cite{fei2021enriching} has compiled large biomedical knowledge graph as well as introduced a separate post-training procedure between pretraining and finetuning where they experimented several strategies using knowledge embedding algorithms to inject domain knowledge. In contrast, our work RadLing attempts to use existing radiology domain-specific knowledge graph RadLex and pretraining objectives that are better suited to the structure of training data.

\section{Methods}

 Our methods consists of preprocessing, tokenization and continuous pretraining. In addition, we have outlined the downstream tasks we have used to evaluate RadLing models. An overview of our approach is presented in Figure \ref{fig:schematic}.

\subsection{Dataset}
We have 545K reports in our training dataset, with over 9.7M sentences and 87M words. The reports have been collected from various medical institutions in the United States and India. We have a high volume of reports for radiograph images taken for head and chest regions (see Appendix Figure \ref{fig:anatomy}). We have a high volume of CT and Xray modalities compared to MRI or Ultrasound (see  Appendix Figure \ref{fig:modality}). 
We preprocessed the dataset using standard techniques, which we describe in more detail in the Appendix \secref{sec:preprocessing}.

\subsection{Tokenization}

We extend existing vocabulary for \Electra so that meaningful tokens in radiology domain are added. We have modified AVocaDo \citep{hong2021avocado} tokenisation to include only domain-specific words in the new vocabulary. We perform Wordpiece tokenization on our corpus to form the set of new tokens $\mathbb{T}_{corpus}$. The vocabulary of BERT contains set of tokens $\mathbb{T}_{BERT}$. We find the tokens that belong to $\mathbb{T}_{corpus} - \mathbb{T}_{BERT}$ and call them  $\mathbb{T}_{candidate}$. We query RadLex to check whether the tokens in $\mathbb{T}_{candidate}$ represent any concept in RadLex. In case the word is present in RadLex, we append the token to the new set of tokens $\mathbb{T}_{new}$.  Since the reports are anonymized, the names, dates and several other patient identifiers are replaced by fixed special tokens, such as `[date]', `[person]', `[location]', `[time]' and `[removed]'. We also include these sets of tokens $\mathbb{T}_s$ into our vocabulary. Our new vocabulary $\mathbb{T}_{RadLing} = \mathbb{T}_{BERT} + \mathbb{T}_{new} + \mathbb{T}_s$.

The addition of significant number of tokens to the existing vocabulary has been seen to introduce catastrophic forgetting or overfitting of model to the new tokens \citep{hong2021avocado}.  We adopt the regularization method in AVocaDo during continuous pretraining to prevent these issues. Total loss during training is a combination of the loss of the pretraining objective and regularization term $\mathcal{L}_{reg}$ (See Appendix \secref{sec:avocado}). 

\subsection{Continuous Pretraining objectives}\seclabel{pretrain-obj}
We train RadLing using \Electra-small \cite{clark2020electra}. \Electra uses generator and discriminator networks to perform `replaced token detection’ (RTD) task. RTD is a pretraining task where the model learns to differentiate between real input tokens and plausible but artificially created replacements. 
The generator is a small MLM model that replaces some corrupted (masked) input tokens by sampling tokens from its vocabulary distribution. 
The discriminator (\Electra) is pretrained to predict whether each token is a replaced token or original token. The advantage of RTD is better language understanding with relatively small amount of pretraining data. 
In our work, we have experimented with three pretraining or self-supervised objectives: Masked language modeling, Section Segmentation, and Knowledge-aware masking. 

\noindent \textbf{Masked Language Modeling (RadLing-MLM):} 
    In keeping with the original paper \cite{devlin2018bert}, for every input sequence, we have randomly masked 15\% of the tokens in the text.    

\noindent \textbf{Section Segmentation (RadLing-SS):} Section segmentation pretraining task follows \cite{kuling2021bi} and uses the discriminator to classify sentences belonging to one of the five report sections: \clinicalsection, \findings, \impressions, \comparison and \miscellaneous \footnote{We use Spacy\citep{honnibal2020spacy} for section segmentation.}. 

\noindent \textbf{Knowledge-aware masking (RadLing-KG):} 
    Knowledge-aware masking utilizes RadLex to intelligently mask (\figref{fig:masking}) 
    so that context information is preserved. RadLex-KG uses four steps to achieve this: 
    \begin{inparaenum}[(a)]
        \item Named entity extraction and entity linking,
        \item Categorization of the entities,
        \item Entity masking according to their categories, and
        \item Regularization during training.
    \end{inparaenum}
    
    From the 15 sub-classes of entities that make up RadLex, we have used the classes "anatomical entity", "clinical finding", "procedure", "imaging observation" and "RadLex descriptors" in this study. Each term in this class can optionally have 63 properties, among which we have only used \textit{Anatomical\_site}.
    
    \noindent \textbf{Named entity extraction and entity linking:} Most of the contextual information of a radiology report is contained in three sections: \clinicalsection, 
    \findings and \impressions. 
    We detect spans of the named entities using SciSpacy \citep{neumann2019scispacy} in these sections of the report. Then, using the NCBO annotator tool\footnote{https://github.com/ncbo/ncbo\_annotator}, we normalize the entities so that we can query RadLex and retrieve information about the ancestors and properties of these entities. 

    \noindent \textbf{Categorization of the entities:}  We classify the entities into three types: \textbf{Symptom}, \textbf{Anatomy} and \textbf{Observation}. Observations refer to different clinical findings in the patient. 
    The normalized entities are classified to these categories using the following heuristics\footnote{There are entities that may have words classified to different categories. For example, `basilar atelectasis' has `basilar' belonging to Anatomies and `atelectasis' belonging to Observations. In these cases we split the entity into its constituents and consider them separately for masking.}:
    \begin{enumerate}
        \item Those that RadLex recognizes as "symptom" are classified as Symptom.
        \item Those that RadLex recognizes as `anatomical entity', `anatomical descriptors' and `anatomically-related descriptor' are classified as Anatomy.
        We also include `location descriptor' class into this category, although they are merely positional qualifiers and don't represent anatomy.
        \item Those that map to the following RadLex sections are classified as Observation:
        `clinical finding', `procedure', `imaging observation', `size descriptor', `normality descriptor', `turbidity descriptor', `stage of healing descriptor' and `composition descriptor'.
    \end{enumerate}

    \noindent \textbf{Entity masking.} We mask entity tokens based on their entity category. We randomly choose one of the following masking options for each sequence:
    \begin{enumerate}
        \item Mask entity tokens identified as Anatomy in the \clinicalsection and \findings of the report. Mask entity tokens identified as Symptom in \findings and \impressions.
        \item Mask entity tokens identified as Observation in the \findings of the report. Mask entity tokens identified as Symptom in \findings and \impressions.
        \item Mask entity tokens identified as Anatomy in the \clinicalsection and \impressions of the report. Mask entity tokens identified as Symptom in \findings and \impressions.
        \item Mask entity tokens identified as Observation in the \impressions of the report. Mask entity tokens identified as Symptom in \findings and \impressions.
    \end{enumerate}
    Each anatomy in a radiology report is associated with a corresponding observation usually. These masking options ensure that context is not masked altogether in a sequence. In addition, if multiple words comprise an entity, we have randomly masked either all or a subset of the tokens comprising the entity. We only consider the masking option that will lead to masking at least 15 \% of the total number of tokens. If no masking option meets that criteria, we randomly select one of the four options and mask the corresponding entities. The rest of the 15\% quota is filled up by tokens corresponding to non-entities.

\noindent \textbf{Regularisation:} In addition to the discriminator loss in \Electra, we introduce a novel regularisation loss where we decrease the penalty for generating a token that belongs to an Observation that pertains to the same Anatomy. This is done by checking the property \textit{Anatomical\_site} of both the linked entity and generated entity or verifying if they belong to the same subclass in `body-system-specific disorder' in RadLex. The regularisation loss is denoted by $\mathcal{L}_{KG}$, and is explained in the Appendix \secref{sec:mask_reg_formula}. 
\section{Experimental Results}
\subsection{Finetuning Tasks}
We evaluated our pretrained models on four finetuning tasks: 
\begin{inparaenum}[(a)]
    \item Named Entity Recognition (NER): extracting radiology-specific anatomies and observations.
    \item Relationship Extraction (RE): extracting relationships between anatomies and observations.
    \item Abnormal classification: classifying reports into normal and abnormal based on the presence or absence of pathologies, and
    \item Radiology Question Answering: providing answers to questions based on radiology reports. 
\end{inparaenum}
We have used RadGraph dataset \citep{jain2021radgraph} for NER and RE tasks, RadQA dataset \citep{soni2022radqa} for Radiology Question Answering, and \cite{demner2016preparing} for abnormal classification. The datasets for these tasks are explained in details in Appendix \secref{sec:finetunedata}.

\begin{table}[t!]
\centering
\footnotesize
\resizebox{.99\linewidth}{!}{
\begin{tabular}{l@{\hspace{0.1cm}}l@{\hspace{0.05cm}}l@{\hspace{0.05cm}}l@{\hspace{0.07cm}}l@{\hspace{0.07cm}}l@{\hspace{0.07cm}}l}
\hline
\textbf{Model} & NER-M & NER-C & RE-M & RE-C & Class & RadQA\\ 
\hline
RadBERT & 0.91 & 0.90 & 0.97 & 0.93 & \textbf{0.99} & \textbf{69.09}  \\
PubMedBERT & 0.86 & 0.89 & 0.78 & 0.69 & 0.98 & 60.08\\ 
\\\hline
RadLing-MLM & 0.89 & 0.89 & 0.98 & 0.94 & 0.98 & 60.96\\
RadLing-SS & 0.89 & 0.88 & 0.96 & 0.94 & 0.97 & 60.23\\
RadLing-KG & \textbf{0.92} & \textbf{0.92} & \textbf{0.98} & \textbf{0.94} & \textbf{0.99} & 62.55\\
 \hline
\end{tabular}
}
\caption{Downstream task Results: For Named Entity Recognition (NER), Relationship Extraction (RE) and Abnormal classification (Class) tasks, macro F1 is reported. For Radiology Question Answering (RadQA), F1 score is reported. NER-M=NER-MIMIC, NER-C=NER-CheXpert, RE-M=RE-MIMIC, RE-C=RE-CheXpert.}
\label{tab:results}
\end{table}

In this section, we present the results from all three RadLing models, which were each fine-tuned on downstream tasks after being pretrained on three different self-supervised objectives (see \secref{pretrain-obj}).

\noindent \textbf{NER.} RadLing-MLM and RadLing-SS do comparatively well on RadGraph test dataset with reference to the state-of-the-art model RadBERT, with F1 scores of 0.89. RadLing-KG however does better than all the other models with F1 score of 0.92. Similar results are seen for CheXpert test dataset, and RadLing-KG performs better than other models with F1 of 0.92. The breakdown of F1 scores in table \ref{table:ner} show that RadLing-KG outperforms the other variants in all the classes, and has significant improvement in the underrepresented class \textit{Observation:Uncertain}.

\noindent \textbf{Relation Extraction.} Both RadLing-MLM and RadLing-KG outperform the state of the art in relationship extraction task, with F1 scores 0.98 and 0.94 on MIMIC and CheXpert test datasets, respectively. Meanwhile, RadLing-SS falls a little short with F1 of 0.96 for MIMIC test dataset. The dataset imbalance does not affect the performance of these models for the underrepresented class [\ref{table:reradgraph}]. Radiologist benchmark macro F1s for this task are 0.91 and 0.704 for MIMIC-CXR and CheXpert respectively, and our models have surpassed it.

\noindent \textbf{Abnormal Classification.} RadLing-KG matches the high performance of RadBERT in this task, with a macro F1 of 0.99, accuracy 99.3\% and AUC of 0.995. 

\noindent \textbf{Question Answering.} For RadQA dataset, RadLing-KG performs the best among the RadLing model variants at F1 of 62.55 and Exact Match of 49.78, but this is lower than the state of the art (69.09 for RadBERT).

\begin{table*}
\footnotesize
\resizebox{0.95\textwidth}{!}{
\begin{tabular}{||c||c|c|c|c||c|c|c|c||}
\hline

      & \multicolumn{4}{c|}{MIMIC} & \multicolumn{4}{c|}{CheXpert} \\

\hline
    \textbf{Model}&\textit{ANAT}&\textit{OBS:DP}&\textit{OBS:U}&\textit{OBS:DA}&\textit{ANAT}&\textit{OBS:DP}&\textit{OBS:U}&\textit{OBS:DA}\\
    \hline
    RadLing-MLM&0.97&0.94&0.75&0.98&0.97&0.98&0.72&0.95\\
    RadLing-SS&0.97&0.94&0.72&0.98&0.97&0.98&0.68&0.95\\
    RadLing-KG&0.98&0.95&0.8&0.98&0.98&0.98&0.81&0.96\\
    \hline
\end{tabular}
}
\caption{Downstream Task Results: Named Entity Recognition (NER) on RadGraph \citep{jain2021radgraph}. Macro F1 scores reported on two test datasets: MIMIC and CheXpert. ANAT refers to Anatomy, OBS refers to Observation, DP: Definitely Present, U: Uncertain, DA: Definitely Absent. }
\label{table:ner}
\end{table*}

\begin{table*}
\footnotesize
\resizebox{0.95\textwidth}{!}{
\begin{tabular}{||c||c|c|c||c|c|c||}
\hline

      & \multicolumn{3}{c|}{MIMIC} & \multicolumn{3}{c|}{CheXpert} \\

\hline
    \textbf{Model}&\textit{Modify}&\textit{Located At}&\textit{Suggestive Of}&\textit{Modify}&\textit{Located At}&\textit{Suggestive Of}\\
    \hline
    RadLing-MLM&0.99&0.98&0.96&0.96&0.93&0.92\\
    RadLing-SS&0.98&0.97&0.94&0.95&0.94&0.9\\
    RadLing-KG&0.99&0.98&0.97&0.96&0.95&0.92\\
    \hline
\end{tabular}
}

\caption{Downstream Task Results: Relation Extraction on RadGraph macro F1 scores. Macro F1 scores reported on two test datasets: MIMIC and CheXpert for 3 relation types: Modify, Located At and Suggestive Of.}
\label{table:reradgraph}
\end{table*}

\subsection{Discussion}
RadLing is one of the first models that has been trained on a radiology report dataset, and the first to use \Electra. RadLing-KG is the first radiology PLM to use RadLex in its training. For this reason, there are not many benchmarks we can compare our results with. Knowledge-aware masking has led to better results in almost all downstream tasks compared to RadBERT. However, RadLing-MLM has comparable performance, and we attribute the robustness to \Electra architecture. The choice of \Electra in our experiments is influenced by its unique architecture that has been shown to yield high performance with low data \cite{clark2020electra}, making it perfect for industry setting. RadLing-KG improves the performance in underrepresented classes for both NER and RE significantly, where all other models perform poorly. To provide an example, uncertain observations comprise only 4.7\% of the NER training data. Now, for a sentence ``mild basilar atelectasis without definite focal consolidation.", ``focal consolidation" is an uncertain observation. Models other than RadLing-KG are not able to capture the whole text as an uncertain observation. Similarly, for relationship extraction task, `suggestive of' reflects 4.7\% of the training data. A sentence like ``Findings are suggestive of mild pulmonary edema with basilar atelectasis" has two `suggestive of' relations : 
\begin{inparaenum}
    \item `Findings' and `edema',
    \item `Findings' and `atelectasis'
\end{inparaenum} 
However, models other than RadLing-KG finetuned for relationship extraction are unable to detect both of these, especially the latter relationship. We surmise that knowledge infusion is a key factor in these stellar results. \Electra has been shown to have a lack of uniformity and alignment where two closely related sentences may have more different representations \cite{meng2021coco}. We hypothesize that this might be one of the main factors contributing to low F1 scores for RadQA. However, we also note that RadLing-KG attempts to counteract this effect and improves on both Exact match and F1.

\section{Conclusion}
In this work we have explored a cost-effective method to train a high performing radiology PLM, RadLing with a small dataset. RadLing models took 2 days to train on Tesla V100 SXM2 machines with 8 GPUs and 16 GB memory per GPU, which using larger models like \Electra-large required 5 days. We developed a knowledge-aware masking strategy to use RadLex to infuse context into radiology PLMs to train RadLing-KG. This led us to the following observations. First, \Electra architecture, without any special pretraining objectives, is able to produce good results with most of the downstream tasks. In a task like relationship extraction, it even outperforms Radgraph radiologist benchmarks. Second, RadLing-KG is the best performing RadLing variant, and outperforms the downstream task benchmarks in all the tasks except QA. Third, Domain specific vocabulary is helpful in better performance of models. In addition, in tasks that use cross attention like vision-language tasks or explainable AI, having unfragmented radiology tokens is helpful. For example, BERT fragments biomedical terms like ‘Thalamus’ into `Tha’, ‘\#\#lam’, ‘\#\#us’, thereby losing the domain-specific meaning, whereas in our work, due to domain specific tokenization, the word 'Thalamus' is retained. Fourth, Infusion of RadLex information  counteracts \Electra limitations in QA dataset. Finally, we have tested the model on proprietary NER datasets and RadLing-KG has yielded 0.92-0.93 macro F1 on less represented anatomies like neck, while for the highly represented anatomies, F1 is as high as 0.98. This actually shows the potential of using RadLex in radiology pretraining. Thus, in a real world setting with high imbalances in datasets, RadLing-KG is more robust.

In future we would like to explore more ways to infuse knowledge by 
\begin{inparaenum}[(a)]
\item Using text description of context like \cite{yuan-etal-2021-improving},
\item Retrieving context from biomedical knowledge graphs like SNOMED \footnote{https://www.nlm.nih.gov/healthit/snomedct/index.html} and UMLS, and
\item more robust knowledge embedding methods.
\end{inparaenum}
We would like to experiment with larger datasets and models, and work with more downstream radiology applications.

\section*{Limitations}
There are a few limitations pertaining to the training data we used. Some of them are listed below.
\begin{enumerate}
    \item RadLing has been trained on English reports only, and therefore will not work out of the box in a multilingual setting.
    \item There is data imbalance with respect to imaging modalities and anatomies covered by our training data. For example, regions like  extremities, neck, spine and shoulder are underrepresented in the dataset, and expected understanding of observations related to those regions may be limited.
    \item There needs to be a study on the diversity of the patients and radiologist expertise represented in the data, and how it impacts the performance of the model for underrepresented communities. 
    \item Different radiologists (and radiology departments) have different preferences and styles of writing reports. In addition, clinical referrals sometimes dictate to what extent some details are documented the report e.g. the Clinical statement. There was no study on the consistency, uncertainty or information richness of the report.
\end{enumerate}
Asides from the training data, there may be space and time throughputs of the model which could make them unsuitable for at-the-edge applications with limited bandwidth.

\section*{Ethics Statement}
The research performed in this paper adheres to the Association for Computing Machinery (ACM) Code of Ethical and Professional Conduct \footnote{https://www.acm.org/code-of-ethics} adopted by the Association for Computational Linguistics (ACL). Radiology reports have been collected from various medical institutions, anonymized and access-controlled to protect PHI information by our dedicated data handling team according to the US Health Insurance Portability Act (HIPAA). To prevent any harm caused due to errors in our model-generated outputs, our models are meant to be deployed in a human-in-the-loop setting where the key information extracted by our models are reviewed by radiologists and physicians.

\section*{Disclaimer}
The concepts and information presented in this paper/presentation are based on research results that are not commercially available. Future commercial availability cannot be guaranteed.

\section*{Acknowledgements}
We extend our sincere gratitude to members in our team for their support to us. We would like to thank The Siemens Big Data team for procuring and guiding us with data handling and anonymization protocols. We would like to thank the various institutions that have provided us with data, including Mt. Sinai and Zwanger-Pesiri Radiology.

\bibliography{anthology,radling}
\bibliographystyle{acl_natbib}

\appendix

\section{Appendix}
\label{sec:appendix}

\begin{figure*}[h!]
\includegraphics[width=\textwidth]{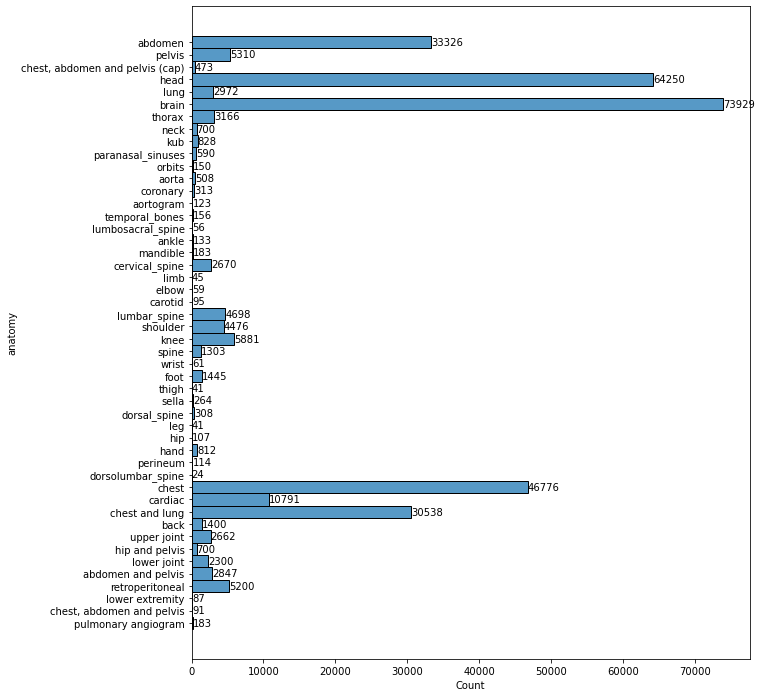}
\centering
\caption{Radiology reports dataset anatomy split}
\label{fig:anatomy}
\end{figure*}
\subsection{Preprocessing}
\seclabel{sec:preprocessing}
Preprocessing of the anonymized radiology reports corpus consists of the following tasks:
\begin{enumerate}[(a)]
\item Regex-based cleaning and normalization: Some of the reports have been converted to text using optical character recognition (OCR). This led to common OCR errors like misspelled character substitutions and insertion of spurious characters. We used manual identification of common errors, followed by Regex-based substitutions for these errors.
\item Section identification: We follow a Regex-based method to split the text in a radiology report into the five sections.
\item Section-based chunking: BERT-like transformers can use maximum 512 tokens as sequence length \cite{michalopoulos-etal-2022-icdbigbird}. We made sure that an entire section of a report is in one section, and divided the report into chunks to fit this restriction.

\end{enumerate}

\subsection{AVocaDo tokenization}
\seclabel{sec:avocado}
In AVocaDo, a contrastive learning framework is employed, and the regularization term $\mathcal{L}_{reg}$ is calculated using the cosine similarity between the sentence encoding outputs from the general PLM and the PLM with a domain-adapted tokenizer as below:

\begin{small}
\begin{equation}
    \begin{split}
         &\mathcal{L}_{reg}(\mathbf{h}_{\mathcal{A}}^{(l)}, \mathbf{h}_{\mathcal{P}}^{(l)}) =  \\
    &\frac{1}{B}\log\sum_{i=1}^{B}\frac{\exp^(sim(\mathbf{h}_{\mathcal{A}, i}^{(l)}, \mathbf{h}_{\mathcal{P}, i}^{(l)})/\tau}{\sum_{j=1}^{B}\exp^(sim(\mathbf{h}_{\mathcal{A}, i}^{(l)}, \mathbf{h}_{\mathcal{P}, j}^{(l)})/\tau}, 
\end{split}
\label{eq:reg}
\end{equation}
\end{small}
where $\mathbf{h}_{\mathcal{A}, i}^{(l)}$ and $\mathbf{h}_{\mathcal{P}, i}^{(l)}$ are l-th layer outputs from the general ($\mathcal{P}$) and adapted ($\mathcal{A}$) PLM encoders for each sentence $x_i$ in a batch of sentences $\mathbf{x}$ of size $B$. $sim(\cdot)$ refers to cosine similarity between the encodings, and $\tau$ is the softmax temperature.

\subsection{Losses in Continuous pretraining objectives}
This section discusses the loss functions we have used for each of our pretraining objectives.
\subsubsection{Masked Language Modeling.}
This objective has the separate losses for the generator and discriminator of \Electra, denoted by $\mathcal{L}_{MLM}(x, \theta_G)$ and $\mathcal{L}_{Disc}(x, \theta_D)$ respectively. They are accompanied by the regularisation term from \ref{eq:reg}, and calculated as follows:
    
   \begin{small}
\begin{equation}
    \begin{split}
         \mathcal{L}_{MLM}(x, \theta_G) & = \lambda_A \mathcal{L}_{reg} + \\ \mathbb{E}\Big(\sum_{i \in m} & - \log p_G(x_i|x^{masked})\Big)\\
    \mathcal{L}_{Disc}(x, \theta_D) & = \lambda_A \mathcal{L}_{reg} + \\ \mathbb{E}\Big(\sum_{t=1}^{n} -\mathbbm{1} & (x^{corrupt}_t = x_t) \log D(x^{corrupt}, t) \\  - \mathbbm{1}( x^{corrupt}_t & \neq x_t) \log(1-D(x^{corrupt}, t)) \Big), 
\end{split}
\label{eq:maskedCE}
\end{equation}
\end{small}
where $p_G$ is the probability of generating a particular token $x_i$ given the masked token $x^{masked}$, $D(\cdot)$ is a sigmoid output of the discriminator that predicts whether the token is ``real", $\lambda_A$ is a regularisation parameter, which is set to 1. 

\subsubsection{Knowledge-Aware Masking}
\seclabel{sec:mask_reg_formula}
\begin{figure}
\includegraphics[width=0.5\textwidth]{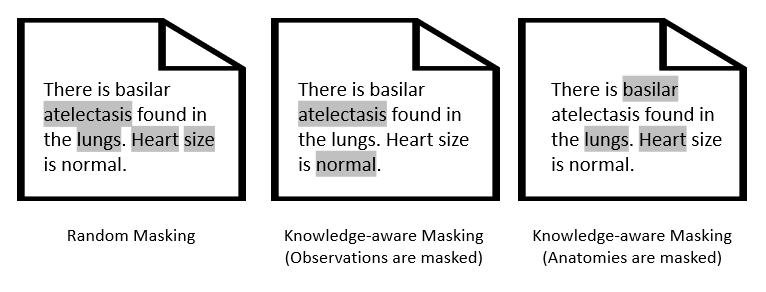}
\centering
\caption{Illustration to show a case where random masking masks all context. Knowledge aware masking masks either anatomy or observation, preserving the context.}
\figlabel{fig:masking}
\end{figure}
The regularisation loss for knowledge-aware masking is denoted by $\mathcal{L}_{KG}$, and calculated as follows:
\begin{small}
\begin{equation}
    \begin{split}
    \mathcal{L}_{KG}(x, \theta_D)  = & \\ \mathbb{E}\Big(\sum_{t=1}^{n}   -\mathbbm{1} & \big(\mathcal{P_A}(x^{corrupt}_t) \in \mathcal{P_A}(x_t)\big) \\ &   \log D(x^{corrupt}, t) \\ - \mathbbm{1} & \big( \mathcal{P_A}(x^{corrupt}_t) \notin \mathcal{P_A}(x_t)\big)\\ &   \log(1-D(x^{corrupt}, t)) \Big),
\end{split}
\label{eq:lossKG}
\end{equation}
\end{small}
where $\mathcal{P_A}(\cdot)$ stands for the anatomical site property of the observation. The final loss for the discriminator is calculated as 
\begin{equation}
    \mathcal{L}_{disc}^{KG} = \mathcal{L}_{Disc} + \lambda_{KG}\mathcal{L}_{KG}
\end{equation}
where $\lambda_{KG}$ is the knowledge graph regularisation parameter, and is set to 1 for our experiments.

\subsection{Experimental setup}
We have trained RadLing using Tesla V100 SXM2 machines, with 8 GPUs and 16 GB memory per GPU. The base model is \Electra-small which has 14M parameters, 12 layers and 256 hidden size. RadLing-MLM was trained in 230 steps, using learning rate 3e-5,  AdamW optimizer and polynomial decay schedule with warmup. RadLing-SS needed 220K steps while RadLing-KG was trained in 235K steps. 

The best performance for NER was achieved after 6 training epochs with learning rate 4e-5, and AdamW optimizer with RadLing-MLM, after 13 steps with RadLing-SS and 9 steps with RadLing-KG. Relationship extraction model was finetuned on RadLing-MLM for 8 epochs with 5e-5 learning rate and AdamW optimizer; 14 epochs with RadLing-SS and 9 with RadLing-KG. Abnormal classification took 19 steps with RadLing-MLM, 10 steps with both RadLing-SS and RadLing-KG with 2e-4 learning rate, dropout 0.2 and AdamW optimizer with 1e-7 epsilon. RadQA finetuning took 8 epochs with 2e-4 learning rate, maximum sequence length 384, document stride 128, maximum query length 128.
All of these models have been trained with early stopping and patience of 3 epochs, and best model selected based on validation loss. For comparison of the results we have used RadBERT-RoBERTa-4m for RadBERT, and the finetuning follows that of RadLing. PubMedBERT results are collected for NER and RE from \cite{jain2021radgraph}. Finetunining for RadQA and Abnormal classification follows finetuning for RadLing and RadBERT.
\begin{figure}
\includegraphics[width=0.5\textwidth]{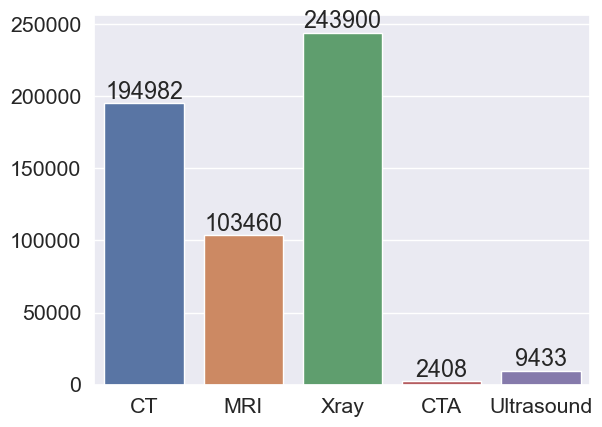}
\centering
\caption{Radiology reports dataset modality split}
\label{fig:modality}
\end{figure}

\subsection{Fine-tuning task details}
\seclabel{sec:finetunedata}

The different finetuning tasks in this paper are described in this section.\\

\textbf{Named Entity Recognition.} The named entity recognition task that we chose to finetune our model on focuses on extracting the anatomy and observations information from the unstructured radiology reports. RadGraph \cite{jain2021radgraph} dataset, annotated by board-certified radiologists, is a collection of annotated training data from MIMIC-CXR dataset and test data from both MIMIC-CXR and CheXpert\cite{irvin2019chexpert} dataset. The anatomies in this schema consists of two concepts: Anatomy and Observation. Anatomy refers to body parts referenced in the radiology reports. Observations refer to the visual findings noted by the radiologist in the medical images (e.g., nodules or opacities),  and pathophysiological processes (diseases) that the radiologist has mentioned (e.g., pneumonia). Observation is further divided into three uncertainty levels: definitely present, uncertain and definitely absent. Training and validation datasets are created by annotating 500 radiology reports from MIMIC-CXR datasets, while test dataset comprises 50 annotated radiology reports from MIMIC-CXR and CheXpert collections. There are 12,388 and 2191 entities in the training and validation datasets respectively. In the test set, there are 1293 and 1473 entities in MIMIC-CXR and CheXpert respectively. Overall, we have an imbalanced dataset with less than 5\% uncertain observations. The evaluation criteria is class-level and aggregate macro F1 and micro F1. Radiologist benchmark macro F1s for this task are 0.981 and 0.894 for MIMIC-CXR and CheXpert respectively.

\textbf{Relationship extraction.} RadGraph dataset also contains relationships between the annotated entities. Relationship extraction refers to the classification of the entity--entity relations given the context of the report. The schema contains three possible relations:  `Suggestive Of', `Located At', and `Modify'. Suggestive Of is a relation between two observations that indicate that the first observation indicates the likelihood or leads to the second one. For example, Streaky densities at the lung base might suggest pneumonia' means the observation `streaky densities' at the lung base indicates that the patient might have pneumonia. Located At is a relation between an observation and anatomy. In the above example, `streaky densities' are 'Located At' the anatomy `lung' base. Modify is a relation between two anatomies or two observations where the first entity is a qualitative or quantitative descriptor of the second entity. There are 9251 relations in the training dataset, 1638 in the validation dataset, 902 in MIMIC-CXR test dataset and 1107 in CheXpert dataset. This dataset is imbalanced with < 6\% of relations being `Suggestive of'. The evaluation metrics for this dataset are macro F1 and micro F1, both on an aggregate level and for each relationship class. Radiologist benchmark macro F1s for this task are 0.91 and 0.704 for MIMIC-CXR and CheXpert respectively.

\textbf{Abnormal classification.} This dataset has been collected from 2 large hospital systems within the Indiana Network for Patient Care database. It contains narrative chest x-ray reports for posterior–anterior (PA) chest x-ray examinations. The dataset\citep{demner2016preparing} contains 3996 de-identified reports, manually and independently annotated by two coders trained in medical informatics. Acute or chronic disease findings, implanted medical devices, or surgical instruments are classified as not normal in this dataset. This dataset also coded in and normalised the abnormalities present in the reports. There are 2564 abnormal reports and the rest are normal. The evaluation metrics for this dataset are macro- and micro-F1 on test dataset.

\textbf{Radiology Question Answering.} This downstream task is an extractive Question Answering application in radiology domain. We use RadQA dataset \citep{soni2022radqa} made from 1009 reports sourced from MIMIC-III- \citep{johnson2016mimic} database, by sampling 100 patients with 1--36 radiology reports. Question creation for this dataset follows a novel approach of basing questions only on clinical referrals of physicians that prompted the radiography being done instead of the whole report. This focuses on the questions that the physicians are the most interested in, as well as the answers to which would most likely be contained in the report. The answer annotations are done by annotators with the full radiology report, with them needing to annotate from Findings and Impression sections. Annotations have been carried out using haystack by expert annotators.  There are 6148 questions in the dataset, among which 1745 are unanswerable. The rest of the answers are extractive and have median and average lengths of 7 and 16.2 respectively. The evaluation of this dataset uses standard Machine Reading Comprehension (MRC) \citep{gardner2019question} metrics, i.e., stricter metric Exact match (EM)  and F1 where word level match is calculated between reference and predicted answers.

\end{document}